\documentclass{article}

\PassOptionsToPackage{numbers, compress}{natbib}
     \usepackage[preprint]{neurips_2020}

\usepackage[utf8]{inputenc} %
\usepackage[T1]{fontenc}    %
\usepackage{hyperref}       %
\usepackage{url}            %
\usepackage{booktabs}       %
\usepackage{amsfonts}       %
\usepackage{nicefrac}       %
\usepackage{microtype}      %

\hypersetup{ %
	pdftitle={Are Few-Shot Learning Benchmarks too Simple ? Solving them without Task Supervision at Test-Time},
	pdfkeywords={},
	pdfborder=0 0 0,
	pdfpagemode=UseNone,
	colorlinks=true,
	linkcolor=blue, %
	citecolor=blue, %
	filecolor=blue, %
	urlcolor=blue, %
	pdfview=FitH,
	pdfauthor={Anonymous},
}

\usepackage{xcolor}
\usepackage{amsmath}
\usepackage{amsfonts}
\usepackage{dsfont}
\usepackage{multirow}
\usepackage{graphicx}
\usepackage{algorithm}
\usepackage{algorithmic}
\usepackage{placeins}
\usepackage{adjustbox}
\usepackage{booktabs}  %

\usepackage{array}
\makeatletter
\newcommand{\thickhline}{%
    \noalign {\ifnum 0=`}\fi \hrule height 1pt
    \futurelet \reserved@a \@xhline
}
\newcolumntype{"}{@{\hskip\tabcolsep\vrule width 1pt\hskip\tabcolsep}}
\makeatother

\newcommand{\X}{\mathcal X}

\newcommand{\Z}{\mathcal Z}

\newcommand{\R}{\mathbb R}

\newcommand\ignore[1]{}
\newcommand{\miniimagenet}{\textit{mini}ImageNet}

\colorlet{lg}{gray!30!white}

\newif\ifintervals
\intervalsfalse
\newcommand{\ci}[1]{\color{gray} \small \ifintervals $\pm$#1 \fi}

\newif\ifadditionaltransductive
\additionaltransductivetrue

\title{Are Few-Shot Learning Benchmarks too Simple ? \\Solving them without Test-Time Labels}

\author{%
  Gabriel Huang\textsuperscript{1}, Hugo Larochelle\textsuperscript{1,2,3} \& Simon Lacoste-Julien\textsuperscript{1,3}\\
  \textsuperscript{1} DIRO, University of Montreal \& Mila, Canada\\
  \textsuperscript{2} Google Brain Montreal, Canada\\
  \textsuperscript{3} Canada CIFAR AI Chair\\
  \texttt{first.last@umontreal.ca} \\
}

\begin{document}

\maketitle

\begin{abstract}
We show that several popular few-shot learning benchmarks can be solved with varying degrees of success without using support set Labels at Test-time (LT). To~this end, we introduce a new baseline called \textit{Centroid Networks}, a modification of Prototypical Networks in which the support set labels are hidden from the method at test-time and have to be recovered through clustering. A benchmark that can be solved perfectly without LT does not require proper task adaptation and is therefore inadequate for evaluating few-shot methods. In practice, most benchmarks cannot be solved perfectly without LT, but running our baseline on any new combinations of architectures and datasets gives insights on the baseline performance to be expected from leveraging a good representation, before any adaptation to the test-time labels.
\end{abstract}

\section{Introduction}

Supervised few-shot classification, sometimes simply called few-shot learning, consists in learning a method that can adapt to new classification tasks from a small number of examples.
Being able to learn new classes from a small number of labeled examples is desirable from a practical perspective because it removes the need for the end-user to label large datasets.
Instead, a central organization with access to large generic datasets could ``pre-train'' the method (training phase) so that when it is shipped to end-users, each user only needs small amounts of labeled data to adapt the method on its own classification task (testing phase).
Supervised few-shot classification is typically formulated as a distribution $P(T)$ of classification tasks, also called episodes, which are split into training, validation, and testing sets. Each episode comes with two small sets of labeled examples called the \textit{support} and \textit{query} sets. The goal is to learn a classifier that can learn (task adaptation) from the task-specific support set $(X_S,Y_S)$ and classify the query set $(X_Q,Y_Q)$ with maximum accuracy, despite limited training data. Typically, this is achieved by training the model on a large number of training episodes beforehand.
The Omniglot~\citep{lake2011one} and \textit{mini}ImageNet \citep{vinyals2016matching} benchmarks have been heavily used to evaluate and compare supervised few-shot classification methods in the last few years~\citep{vinyals2016matching,ravi2016optimization, snell2017prototypical,finn2017model,sung2018learning}. 

An important question is whether those benchmarks are actually evaluating the task adaptation capabilities of few-shot methods, or rather something else. 
Consider the following (transductive) classification task, extracted from the ``Mongolian'' alphabet of Omniglot: %
\begin{center}
\includegraphics[width=0.5\linewidth]{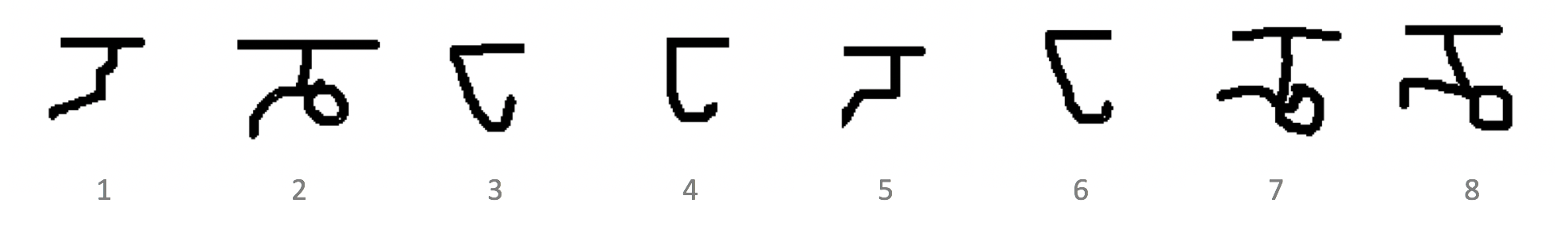}
\end{center}
It is relatively easy to solve the task even without being familiar with ``Mongolian'' characters because we are already familiar with the ``group by character'' task.\footnote{Solution: there are 3 classes $\lbrace 1,5\rbrace, \lbrace 2,7,8\rbrace, \lbrace 3,4,6\rbrace$.} We can solve the task without receiving a single labeled example of the new classes, that is, without using a support set.
Conversely, there are task distributions that do require learning from a support set.
For instance, consider this other transductive classification task : %
\begin{center}
\includegraphics[width=0.5\linewidth]{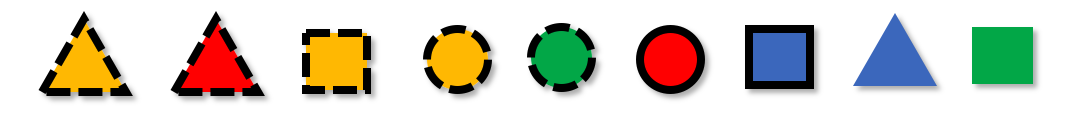}
\end{center}
The task is now ambiguous because there are many possible semantics (by shape, color, or border style). 
In this case, we would need a support set to learn (task adaptation) which criterion should be used.
Proper task adaptation requires using the support set labels at test-time. Therefore, one way to investigate which few-shot benchmarks require proper task adaptation is to ask \textit{which benchmarks can be solved without using support set labels at test-time ?}

\textbf{Definition (LT vs. NLT methods).}
\textit{
We say that a few-shot classification method is \textbf{LT} if it uses the support set \textbf{L}abels at \textbf{T}est-time.
Conversely, we say that a few-shot classification method is \textbf{NLT} if it uses \textbf{N}o \textbf{L}abels at \textbf{T}est-time. Both LT and NLT methods are allowed to use labels during training. Note that all usual few-shot methods~\citep{vinyals2016matching,finn2017model,snell2017prototypical} are LT methods}

Given this definition, we want to know if we can reach high performance on a benchmark with NLT methods. 
In particular, benchmarks that can be solved by NLT methods are definitely too ``simple'' and do not require proper task adaptation, in the sense of using test-time labels.
We have reasons to suspect that NLT methods can reach high performance on Omniglot and miniImageNet because their class semantics are invariant across different episodes (Omniglot classes are always alphabet characters, \textit{mini}ImageNet classes are always object categories as defined by the WordNet taxonomy~\citep{miller1995wordnet,russakovsky2015imagenet}), and because the classes are randomly split between training and testing (there is no domain shift).

We introduce a new baseline called \textit{Centroid Networks} (or CentroidNet), which is an NLT modification of Prototypical Networks in which the support set labels are hidden from the method and recovered through clustering. 
We attempt to solve several popular few-shot benchmarks using our NLT baseline and report the resulting accuracies. Those numbers can be interpreted as \textit{the baseline performance to be expected from leveraging a good representation, before any LT-based task adaptation takes place}. Therefore, we recommend running our NLT baseline on any published combinations of architectures and datasets in order to disentangle the fraction of the performance can be attributed to the architecture/representation versus its ability to learn from new labels.

Our contributions are the following :
\begin{itemize}
    \item We propose a simple NLT baseline for solving few-shot tasks without support set labels at test-time. This baseline helps determine how much performance is to be expected from leveraging a good representation without learning from task-specific labels at test-time.
    \item We show that Centroid Networks achieve 99.1\% and 98.1\% NLT accuracies for Omniglot 5-way/5-shot and 20-way/5-shot, effectively showing that Omniglot can be solved without~LT.
    \item We report NLT accuracies of our baseline on several other popular benchmarks, giving an idea of the performance to be expected on any (dataset, architecture) combination from leveraging a good representation before any LT-based task adaptation takes place. 
    We~observe that support set labels are much more critical for cross-domain benchmarks.
\end{itemize}

Finally, we also explore applying our method to few-shot clustering and (LT) transductive few-shot learning.

\section{Related Work}

\textbf{Simple Baselines.} \citet{chen2019closer} propose a simple baseline for LT few-shot classification, and show that even simple baselines can solve few-shot benchmarks with good performance when combined with the right architecture.
Our NLT baseline is even simpler in the sense that we neither use LT or change the representation at test-time.

\textbf{Task Overfitting vs. Task Generalization.} \citet{yin2019meta} introduce the notion of non-mutually exclusive tasks, which is a family of tasks that can be solved by a single function. They argue that non-mutually exclusive tasks are not diverse enough and that few-shot classification methods are at risk of ``task overfitting''. They propose a regularization technique to prevent task overfitting. In our case, we argue that even the original (mutually-exclusive) Omniglot and \textit{mini}ImageNet do not have sufficient task diversity in terms of class semantics. Our motivation is different and is to show that one can memorize the class semantics of the training set and still achieve high accuracy on the test set, thereby making those benchmarks too ``simple''.

\textbf{Task Adaptation vs. Feature Reuse.} \citet{raghu2019rapid} investigate whether the performance of MAML~\citep{finn2017model} comes from rapid adaptation to new tasks or from reusing good features. They propose a version of MAML in which the features are kept constant in the inner loop, without any loss in performance. Our work shares the same motivation in showing that the performance of meta-learning algorithms might come more from using good universal features rather than doing task adaptation, where we mean ``without using LT'' and they mean ``withot adapting the features''. We take it one step further than fixing the features by proposing a NLT baseline.

\textbf{Meta-Learning and Semi/Un-supervised Learning.} 
Some recent work has explored combinations of unsupervised learning and meta-learning, to address various other tasks. \citet{metz2018learning} meta-learn an unsupervised representation update rule that produces useful features for LT few-shot learning. 
Similarly, \citet{hsu2018unsupervised,khodadadeh2018unsupervised} learn a LT method using no labels a training time.
Thus, all these works consider the opposite setting from us : they use no labels at training-time to learn a LT method, while we use labels at training time to learn a NLT method.
Our work is also related to Semi-Supervised Prototypical Networks~\citep{ren2018meta}, in which the support set contains both labeled and unlabeled examples. In a sense, we go beyond their work by requiring no labels at all to infer centroids at test-time.

\textbf{Zero-Shot Learning.} Strictly speaking, \textit{true} zero-shot learning would mean solving the few-shot classification problem without using the support set at all.
Centroid Networks are related to true zero-shot learning because they do not use the labels of the support set. However, our method uses the images and some partial information (Section~\ref{sec:taskseval}) and therefore cannot be considered pure zero-shot learning. 
In practice, zero-shot methods also have access to partial information in the form of text descriptions of the new classes~\citep{socher2013zero,romera2015embarrassingly}.

Our work is also related to the supervised clustering, learning to cluster, and other clustering-related literature. We have moved them to Appendix~\ref{sec:additionalrelatedwork} due to a lack of space and because learning-to-cluster is not the central contribution of this paper.

\section{Few-shot Tasks and Evaluation\label{sec:taskseval}}

We give some background on meta-learning and standard (LT) few-shot classification. Then, we will introduce NLT few-shot classification, which formalizes what it means to \textit{solve few-shot classification without test-time labels}. 

Meta-learning consists in learning a method that can solve a distribution of tasks or episodes $P(T)$ well, with respect to some external evaluation metric (e.g. accuracy).
Typically, independently sampled tasks from $P(T)$ are split into training, validation, and testing sets, although the meta-learning tasks generally contain additional supervision as we discuss below.
In this framework, an algorithm is first trained to solve the training tasks (train-time), then it is evaluated on the validation/testing tasks (test-time).

\paragraph{LT few-shot classification.} Each episode comes with a small \textit{support} set $S=(X_S,Y_S)$ and a small \textit{query} set $Q=(X_Q,Y_Q)$, where $X_S,X_Q$ denote images or data, and $Y_S,Y_Q$ their labels. 
The task is to predict labels $\widehat Y_Q$ for the query images $X_Q$ and the learner has access to the task-specific images $X_S$ and labels $Y_S$ for task adaptation. Finally, the classification accuracy is computed between $\widehat Y_Q$ and $Y_Q$. We denote it \textbf{LT-accuracy} to differentiate it from the metrics of other tasks.

\paragraph{NLT few-shot classification.} The NLT setting is the same as its LT counterpart except that we hide the support set labels at test-time.
Before making predictions, we need to introduce an additional step in which the learner attempts to recover $Y_S$ by clustering $X_S$ into $K$ clusters. Denote $\widehat C_{S}^{(i)} \in [1,K]$ the cluster index for the $i$-th example $X_S^{(i)}$. Because the cluster indices are only defined up to permutation of the values, we use the Hungarian algorithm\footnote{\url{https://docs.scipy.org/doc/scipy/reference/generated/scipy.optimize.linear_sum_assignment.html}} to find the permutation $\sigma: [1,K] \rightarrow [1,K]$ that maximizes the accuracy $\sum_i \mathbf{1}_{\lbrace \sigma(\widehat C_S^{(i)}) \neq Y_S^{(i)} \rbrace}$ on the support set, and denote $\widehat Y_{S}^{(i)} = \sigma(Y_{S}^{(i)})$ the optimal indices.
The final task is to predict labels $\widehat Y_Q$ for $X_Q$ as usual, after learning from the reconstructed support set $(X_S,\widehat Y_S)$. The predictions $\widehat Y_Q$ are compared with the ground-truth $Y_Q$. We denote the resulting accuracy \textbf{NLT-accuracy}. Note that any LT-method can be made into a NLT-method by combining it with a clustering algorithm. This is exactly the approach of our NLT baseline, in which ProtoNet is combined with Sinkhorn K-means.

Note that this section is about the evaluation tasks  (the testing tasks) and does not put restrictions on the usable labels during training. 
Typically, standard LT few-shot classification methods have access to $Y_Q$ during training but $Y_Q$ is reserved for external evaluation during testing.
In the same spirit, NLT few-shot classification methods are allowed to use $Y_S$ and $Y_Q$ during training, but both $Y_S$ and $Y_Q$ are reserved for external evaluation at testing time.

\section{Centroid Networks\label{sec:centroidnet}}

We mentioned previously that any LT method can be made into a NLT method by combining it with a clustering step at test-time (Section~\ref{sec:taskseval}). Our NLT baseline, \textit{Centroid Networks} or \textit{CentroidNet}, is a combination of ProtoNet and Sinkhorn K-Means, which is used to recover the support set labels at test-time.

%

\subsection{Prototypical Networks}

Prototypical Networks or ProtoNets~\citep{snell2017prototypical} is one of the simplest supervised few-shot classification methods, and yet it has been shown to be competitive with more complex methods when combined with ResNet backbones~\citep{ye2020fewshot}. The only learnable component of ProtoNets is the (backbone) embedding function $h_\theta:\X\rightarrow \Z$ which maps images to an embedding (feature) space. Given an episode,  ProtoNet takes the support set images, maps them to the feature space, and computes the average (the prototype) of each class $\mu_j = \frac 1 M \sum_i h_\theta(x_i^s) * \mathds 1\lbrace y_i^s=j\rbrace$ . Each point from the query set is then classified according to a soft nearest-neighbor scheme $p_\theta(y=j|x) = \textrm{softmax}( \-||h_\theta(x) - \mu_j||^2 )$. ProtoNets are trained end-to-end by minimizing the classification cross-entropy on the query set.

\subsection{Sinkhorn K-Means\label{sec:sinkhornkmeans}}

We propose \textit{Sinkhorn K-Means} as the clustering module of Centroid Networks. The idea of Sinkhorn K-Means has been mentioned sporadically in the literature (see Appendix~\ref{sec:additionalrelatedwork}) but to the best of our knowledge the algorithm has never been explicitly described or applied to few-shot learning before.

\newcommand{\comment}[1]{\quad {\color{blue!50!black}// #1}}

\begin{figure}
\centering
\begin{minipage}{0.53\textwidth}
\begin{algorithm}[H]
   \caption{$\textrm{Sinkhorn}(x,c,\gamma)$ for empirical distributions.}
   \label{alg:sinkhorn}
\begin{algorithmic}
   \STATE {\bfseries Input:} data $(x_i)_{1\leq i\leq n} \in\R^{n\times d}$, centroids $(c_j)_{1\leq j\leq k} \in\R^{k\times d}$, regularization constant $\gamma>0$.
   \STATE {\bfseries Output:} optimal transport plan $(p_{i,j})\in \R^{n\times k}$.
   \STATE $ K_{i,j} \leftarrow \exp(-||x_i-c_j||_2^2 / \gamma)$ $\in\R^{n\times d}$
     \STATE $R_i \leftarrow 1/n \color{lg} \quad 1\leq i \leq n$ 
   \STATE $v_j \leftarrow 1, , C_j \leftarrow 1/k \quad \color{lg} 1\leq j \leq d $  
   \WHILE{not converged}
   \STATE
   $ u_i \leftarrow R_i / (\sum_{j=1}^k K_{i,j} v_j ), \color{lg}\quad 1\leq i \leq n $   
   \STATE 
   $ v_j \leftarrow C_j / (\sum_{i=1}^n K_{i,j} u_i ),  \color{lg}\quad 1\leq j  \leq k $ 
   \ENDWHILE
   \STATE $ p_{i,j} \leftarrow u_i K_{i,j} v_j, \quad\color{lg} 1\leq i\leq n, 1\leq j\leq k $ 
   \STATE \textbf{return }{assignments $p$}
\end{algorithmic}
\end{algorithm}
\end{minipage}
\hfill
\begin{minipage}{0.43\textwidth}
\begin{algorithm}[H]
   \caption{$\textrm{Sinkhorn K-Means}(x,c,\gamma)$}
   \label{alg:wassersteinkmeans}
\begin{algorithmic}
   \STATE {\bfseries Input:} data $(x_i)_{1\leq i\leq N}$, initial centroids $(c_j)_{1\leq j\leq K}$, regularization constant $\gamma > 0$.
   \STATE {\bfseries Output:} final centroids $(c_j)_{1\leq j\leq K}$, optimal assignment $(p_{i,j}) \in\R^{N\times K}$.
   \WHILE{not converged}
   \STATE    $(p_{i,j}) \leftarrow \textrm{Sinkhorn}(x, c, \gamma)$ 
   \STATE    $c_j \leftarrow k \sum_{i=1}^n p_{i,j} x_i, \quad \color{lg} 1\leq j\leq k$
   \ENDWHILE
   \STATE {\textbf{return} centroids $c$, assignments $p$.}
\end{algorithmic}
\end{algorithm}
\end{minipage}
\end{figure}

Sinkhorn K-Means takes as input a set of $N$ points $x\in\R^{n\times d}$ (in our case, ProtoNet embeddings) and outputs a set of $K$ centroids $c_j\in\R^{k\times d}$ with data-centroid soft assignments $p\in\R^{n\times k}$. We start by initializing the centroids around zero with a small amount of Gaussian noise to break symmetries. Then, we attempt to find the centroids that minimize the Sinkhorn distance~\citep{cuturi2014fast} between the empirical distributions defined by the data $p(x) = \frac{1}{N}\sum_{i=1}^N \delta(x-x_i)$ and the centroids $q(x) = \frac{1}{K}\sum_{j=1}^k \delta(x-c_j)$. To do so, we alternatively compute the optimal transport plan between $p$ and $q$ using Sinkhorn distances (Algorithm~\ref{alg:sinkhorn}) and update each centroid to the weighted average of its assigned data points. For simplicity, Algorithm~\ref{alg:wassersteinkmeans} describes the procedure in the case where clusters are balanced. When the clusters are not balanced but the cluster weights are known (e.g.\ Meta-Dataset), the weights can be taken into account by the Sinkhorn distance. All details can be found in our code.  

Sinkhorn K-Means can be seen as a version of K-Means where the greedy nearest-centroid hard assignment (expectation step) is replaced with a global regularized optimal transport soft-assignment. More discussions about the Sinkhorn distance, the differences between Sinkhorn/Regular K-Means, and an empirical comparison are given in Sections~\ref{app:sinkhorn} and~\ref{sec:ablation} of the Appendix. 

\subsection{Combining ProtoNet with Sinkhorn K-Means}

For training, we teach ProtoNet to solve LT few-shot classification tasks (regular ProtoNet training).
For testing, we combined Sinkhorn K-Means as described below.

\paragraph{Training (all tasks).} Training is the same regardless of the evaluation task. Using standard training~\citep{snell2017prototypical}, we fit the ProtoNet backbone to solve LT few-shot classification.
Given a training episode, we embed the support set, compute the prototypes, make predictions on the query set, compute the cross-entropy loss after revealing the query labels, and minimize it with gradient descent with respected to the parameters of the backbone.
In some cases, we have found helpful to use an additional center loss~\citep{wen2016discriminative} in order to pull the embeddings of the same class together. This is discussed in the ablation study in Appendix~\ref{sec:ablation}.

\paragraph{Testing on NLT few-shot classification.} Given the support and query images $X_S,X_Q$, the number of classes $K$ and number of shots, we embed the support images and compute centroids $(c_j)$ and the optimal transport plan $(p_{i,j})$ using Sinkhorn K-Means. To get hard assignments, we can either classify query points according to their nearest centroid (\textit{softmax assignments}), or return their majority assigned centroid $\arg\!\max_j p_{i,j}$ (\textit{sinkhorn assignments}). According to the ablation study, the choice of assignment strategy has little effect on the performance (Appendix~\ref{sec:ablation}).

\section{Experiments}

We start with our main results for NLT few-shot classification (Section~\ref{sec:expcrosstask}). We then apply our method to cross-domain benchmarks, which we expect to be harder without test-time labels (Section~\ref{sec:expcrossdomain}). We also explore applying our method in other few-shot settings such as few-shot clustering with partial information and transductive few-shot classification (Section~\ref{sec:expother}).
Details about the implementation and datasets can be found in Appendix~\ref{sec:implementationdetails}.

\subsection{Main Results\label{sec:expcrosstask}}

\textbf{[Table~\ref{table:mainresult}]} We run our NLT baseline on four popular few-shot classification benchmarks: Omniglot~\citep{lake2011one}, \textit{mini}ImageNet~\citep{vinyals2016matching}, \textit{tiered}ImageNet~\citep{ren2018meta}, and CUB~\citep{wah2011caltech}.
We consider the classic four layer convolutional architecture~\citep{snell2017prototypical} (which we denote Conv), and the ResNet-12~\citep{ye2020fewshot}. 
We combine Sinkhorn K-Means with the ProtoNet implementation of~\citep{ye2020fewshot}, except for Omniglot where we use the original implementation~\citep{snell2017prototypical}.
We report NLT accuracies alongside several state-of-the art LT methods for comparison.
Given the very high numbers for Omniglot, we can conclude that both the 5-way 5-shot and 20 way 5-shot settings can be solved without LT. 
The same type of definitive conclusions cannot be made about the other bencharks, but we do observe that for \textit{tiered}ImageNet our NLT baseline surpasses a number of LT methods implemented in a recent survey~\citep{chen2019closer}.

\begin{table}[]
\centering
\caption{LT and NLT few-shot classification on same-domain benchmarks. We consider 5-way 5-shot episodes, plus an additional 20-way 5-shot for Omniglot. Test accuracies are computed over 600 test episodes. The ResNet architectures are ResNet-12 except for SimpleShot and CTM which use ResNet-18. We denote ProtoNet* the implementation from which we derive CentroidNet. For the CUB dataset, some results~\citep{ye2020fewshot} that we could not reproduce are in gray (they might have accidentally reported validation accuracies).\label{table:mainresult}
}

\newcommand{\tableratio}{0.4}

\begin{adjustbox}{width=\tableratio\linewidth}
\begin{tabular}{lcc}
\toprule
\multicolumn{3}{c}{\textit{mini}ImageNet}\\
\midrule
\textbf{LT Methods} & \multicolumn{2}{c}{\textbf{LT Accuracy}} \\
\textcolor{lg}{backbone $\longrightarrow$}  & Conv & ResNet\\
\midrule
MatchNet~\citep{vinyals2016matching} & 51.09\ci{0.71} & -\\
MAML~\citep{finn2017model} & 63.11\ci{0.92}  & -\\
RelationNet~\citep{sung2018learning}  & 67.07\ci{0.69}  & -\\
ProtoNet~\citep{snell2017prototypical} &  68.20\ci{0.66}  & -\\
FEAT~\citep{ye2020fewshot} &  71.61\ci{0.16}  & -\\
TADAM~\citep{oreshkin2018tadam}  & -& 76.70\ci{0.30}\\
MetaOptNet~\citep{lee2019meta}   & -& 78.63\ci{0.46}\\
SimpleShot~\citep{wang2019simpleshot}   & -& 80.02\ci{0.14}\\
CTM~\citep{li2019finding}  & -& 80.51\ci{0.13}\\
ProtoNet*~\citep{ye2020fewshot}  & 71.33\ci{0.16} & 80.53\ci{0.14}\\
FEAT~\citep{ye2020fewshot}   & -& 82.05\ci{0.14}\\
\midrule
\textbf{NLT Methods} & \multicolumn{2}{c}{\textbf{NLT Accuracy}} \\
\textcolor{lg}{backbone $\longrightarrow$}  & Conv & ResNet\\
\midrule
CentroidNet (ours) &  57.57\ci{0.94} & 69.86\ci{0.94}\\
\bottomrule
\end{tabular}
\end{adjustbox}
\quad\quad
\begin{adjustbox}{width=\tableratio\linewidth}
\begin{tabular}{lc}
\toprule
\multicolumn{2}{c}{\textit{tiered}ImageNet}\\
\midrule
\textbf{LT Methods} & \multicolumn{1}{c}{\textbf{LT Accuracy}} \\
\textcolor{lg}{backbone $\longrightarrow$}  & ResNet\\
\midrule
ProtoNet~\citep{snell2017prototypical} &  72.69\ci{0.74}\\
RelationNet~\citep{sung2018learning}  & 71.32\ci{0.78}\\
MetaOptNet~\citep{lee2019meta}   & 81.56\ci{0.63}\\
CTM~\citep{li2019finding}  & 84.28\ci{1.73}\\
SimpleShot~\citep{wang2019simpleshot} & 84.58\ci{0.16}\\
ProtoNet*~\citep{ye2020fewshot} &  84.03\ci{0.16}\\
FEAT~\citep{ye2020fewshot} &  84.79\ci{0.16}\\
\midrule
\textbf{NLT Methods} & \multicolumn{1}{c}{\textbf{NLT Accuracy}} \\
\textcolor{lg}{backbone $\longrightarrow$}  & ResNet\\
\midrule
CentroidNet (ours) &  75.36\ci{1.04}\\
\thickhline
\end{tabular}
\end{adjustbox}
\begin{adjustbox}{width=\tableratio\linewidth}
\begin{tabular}{lcc}
\toprule
\multicolumn{3}{c}{Omniglot}\\
\midrule
\textbf{LT Methods} & \textbf{5-way} & \textbf{20-way}\\
\textcolor{lg}{backbone $\longrightarrow$}  &  \multicolumn{2}{c}{ConvNet}\\
\midrule
SiameseNet~\citep{koch2015siamese} & 98.4 & 97.0\\
MatchNet~\citep{vinyals2016matching} & 98.9 & 98.5\\
NeuralStat~\citep{edwards2016towards} & 99.5 & 98.1\\
MemoryMod~\citep{kaiser2016learning} &99.6 & 98.6\\
ProtoNet*~\citep{snell2017prototypical} &  99.7 & 98.9\\
MAML~\citep{finn2017model} & 99.9 & 98.9 \\
\midrule
\textbf{NLT Methods} & \textbf{5-way} & \textbf{20-way}\\
\textcolor{lg}{backbone $\longrightarrow$}  & \multicolumn{2}{c}{ConvNet}\\
\midrule
CentroidNet (ours) &  99.1\ci{0.1} & 98.1\ci{0.1}\\
\bottomrule
\end{tabular}
\end{adjustbox}
\quad\quad
\begin{adjustbox}{width=\tableratio\linewidth}
\begin{tabular}{lc}
\toprule
\multicolumn{2}{c}{CUB}\\
\midrule
\textbf{LT Methods} & \multicolumn{1}{c}{\textbf{LT Accuracy}} \\
\textcolor{lg}{backbone $\longrightarrow$}  & ConvNet\\
\hline
\addlinespace[0.2em]
MatchNet~\citep{vinyals2016matching} &  72.86\ci{0.70}\\
MAML~\citep{finn2017model} &  72.09\ci{0.76}\\
ProtoNet~\citep{snell2017prototypical} &  70.77\ci{0.69}\\
ProtoNet* (repro)~\citep{ye2020fewshot} &  75.33\ci{0.71}\\
RelationNet~\citep{sung2018learning}  & 76.11\ci{0.69}\\
\color{lg} MatchNet~\citep{ye2020fewshot} & \color{lg} 79.00\ci{0.16}\\
\color{lg} ProtoNet~\citep{ye2020fewshot} & \color{lg} 81.50\ci{0.15}\\
\color{lg} FEAT~\citep{ye2020fewshot} & \color{lg} 82.90\ci{0.15}\\
\midrule
\textbf{NLT Methods} & \multicolumn{1}{c}{\textbf{NLT Accuracy}} \\
\textcolor{lg}{backbone $\longrightarrow$}  & ConvNet\\
\midrule
CentroidNet (ours) &  66.13\ci{1.08}\\
\bottomrule
\end{tabular}
\end{adjustbox}

\end{table}

\subsection{Cross-Domain NLT Few-Shot Classification \label{sec:expcrossdomain}}

\begin{table}
\large
\centering
\caption{LT and NLT few-shot classification on cross-domain benchmarks. \textit{mini}ImageNet$\rightarrow$CUB is 5-way 5-shot, and Meta-Dataset involves variable numbers of ways and shots~\citep{triantafillou2019meta}.
Test accuracies are averaged over 600 episodes. See Appendix~\ref{sec:expmetadatasetci} for confidence intervals.\label{table:crossdomain}}

\centerline{
\begin{minipage}{1.0\textwidth}
\begin{adjustbox}{width=0.8\textwidth}
\begin{tabular}{l | ccc|c || ccc|c}
\toprule
\multicolumn{9}{c}{Meta-Dataset}\\
\midrule
& \multicolumn{4}{c||}{\textit{Train on ILSVRC}}
& \multicolumn{4}{c}{\textit{Train on all datasets}}\\
\textbf{Test Dataset} & \multicolumn{3}{c|}{\textbf{LT}} & \textbf{NLT}
& \multicolumn{3}{c|}{\textbf{LT}} & \textbf{NLT}\\
\color{lg} method $\rightarrow$
& \multicolumn{1}{c}{Proto} & CNAPs  & \multicolumn{1}{c|}{SUR} & Centro
& \multicolumn{1}{c}{Proto}  & CNAPs & \multicolumn{1}{c|}{SUR} & Centro \\
\midrule
ILSVRC & 
44.12  &    50.6 &  56.3   &\text{26.40}\ci{0.88}  & 
41.79  &    52.3 &  56.3   &\text{23.84}\ci{0.82}   \\
Omniglot &          
53.40  &    45.2 &  67.5   &\text{36.83}\ci{1.20}  & 
81.93  &    88.4 &  93.1   &\text{66.25}\ci{1.12}   \\
Aircraft &          
45.29  &    36.0 &  50.4   &\text{24.15}\ci{0.72}  &  
69.43  &    80.5 &  85.4   &\text{57.50}\ci{1.01}   \\
Birds &             
63.59  &    60.7 &  71.7   &\text{41.08}\ci{1.05}  & 
64.73  &    72.2 &  71.4   &\text{43.56}\ci{1.03}   \\
Textures &          
61.78  &    67.5 &  70.2  &\text{39.63}\ci{0.70}  & 
66.35  &    58.3 &  71.5   &\text{43.50}\ci{0.76}   \\
QuickDraw &         
49.58  &    42.3 &  52.4   &\text{31.04}\ci{0.95}  & 
67.74  &    72.5 &  81.3   &\text{46.96}\ci{1.04}   \\
Fungi &             
35.27  &    30.1 &  39.1   &\text{18.11}\ci{0.71}  &
38.94  &    47.4 &  63.1   &\text{21.76}\ci{0.76}   \\
VGG Flower &         
78.09  &    70.7 &  84.3   &\text{47.98}\ci{0.96} & 
84.45  &    86.0 &  82.8   &\text{55.11}\ci{0.95}   \\
\midrule            
Traffic Sign         & 
46.08  &    53.3 &  63.1   &\text{22.03}\ci{0.66}  & 
49.91  &    60.2 &  70.4   &\text{22.71}\ci{0.66}   \\
MSCOCO &            
35.63  &    45.2 &  52.8   &\text{18.19}\ci{0.69}  & 
36.64  &    42.6 &  52.4   &\text{17.60}\ci{0.77}   \\
\bottomrule
\end{tabular}
\end{adjustbox}
\newline
\vspace*{1em}
\newline
\centering 
\begin{adjustbox}{width=0.4\textwidth}
\begin{tabular}{lcc}
\toprule
\multicolumn{3}{c}{MiniImageNet $\longrightarrow$ CUB}\\
\midrule
\textbf{LT Methods} & \multicolumn{2}{c}{\textbf{LT Accuracy}} \\
\textcolor{lg}{backbone $\longrightarrow$}  & ConvNet & ResNet\\
\midrule
MAML~\citep{chen2019closer} & - & 51.34\ci{0.72} \\
MatchNet~\citep{chen2019closer} & - & 53.07\ci{0.74} \\
RelationNet~\citep{chen2019closer} & - & 57.71\ci{0.73} \\
ProtoNet* (repro) & 62.52\ci{0.73} & 61.38\ci{0.76} \\
ProtoNet~\citep{chen2019closer} & - & 62.02\ci{0.70} \\
Baseline~\citep{chen2019closer} & - & 65.57\ci{0.70}\\
GNN-FT~\citep{tseng2020cross} & - & 66.32\ci{0.80}\\
Neg-Softmax~\citep{liu2020negative} & - & 69.30\ci{0.73} \\
\midrule
\textbf{NLT Methods} & \multicolumn{2}{c}{\textbf{NLT Accuracy}} \\
\textcolor{lg}{backbone $\longrightarrow$}  & ConvNet & ResNet\\
\midrule
CentroidNet (ours) & 47.01\ci{0.91} & 44.62\ci{0.90} \\
\bottomrule
\end{tabular}
\end{adjustbox}
\end{minipage}
}
\end{table}

\textbf{[Table~\ref{table:crossdomain}]} 
We run our NLT baseline on two cross-domain benchmarks, \textit{mini}ImageNet$\rightarrow$CUB~\citep{chen2019closer} and Meta-Dataset~\citep{triantafillou2019meta} which was recently proposed as a harder benchmark.
We report NLT accuracies alongside several LT methods, including the ProtoNet implementations that we are based on~\citep{triantafillou2019meta,ye2020fewshot} and recent state-of-the-art LT-methods on Meta-Dataset :  CNAPs~\citep{requeima2019fast} and SUR~\citep{dvornik2020selecting}.
For both cross-domain benchmarks, we observe that the gap between our NLT baseline and the LT state-of-the-art is significantly larger than in the same-domain case, thereby confirming that cross-domain benchmarks are much more dependent on using test-time labels, which would make them more appropriate benchmarks for validating the task-adaptation capabilities of LT methods.

\subsection{Exploring Other Few-Shot Settings \label{sec:expother}}

\intervalstrue

\begin{table}
\caption{Few-shot clustering on Omniglot with CCN splits.  We consider the usual 4-layer ``Conv-4'' architecture~\citep{snell2017prototypical}, the ``CCN'' architecture used by~\citet{hsu2017learning} which has more filters, and using raw images ($32\times 32$). The differences between the two architectures are not significant. Our results are reported with 95\% confidence intervals and averaged over 1000 test episodes with a fixed model. Numbers with a star* are from~\cite{hsu2018multiclass}.\label{table:omniglot_ccn}}
\begin{center}
 \begin{tabular}{ l l l l}
\toprule
\multicolumn{4}{c}{Omniglot-CCN} \\
\midrule
\textbf{Clustering Methods} & \multicolumn{3}{c}{\textbf{Clustering Accuracy}}\\ 
\textcolor{lg}{backbone $\longrightarrow$}  & $32\times 32$ & Conv-4 & CCN\\
\midrule
K-Means  & 21.7* & 69.4\ci{0.5} & -   \\
CCN (KCL)~\citep{hsu2017learning}  & - & - & 82.4*  \\
CCN (MCL)~\citep{hsu2018multiclass}  & - & - & 83.3*  \\
CentroidNet-FSC (ours) & - & \textbf{86.8}{\color{lg} \ci{0.6}} & \textbf{86.6}{\color{lg} \ci{0.6}}\\
\bottomrule
\end{tabular}
\end{center}
\end{table}

\paragraph{Few-Shot Clustering with Partial Information [Table~\ref{table:omniglot_ccn}].}
In few-shot clustering, the goal is to cluster new sets of data according to semantics learned during training. We investigate using CentroidNet for few-shot clustering. Please note that our method is less flexible than other learning to cluster approaches~\citep{hsu2018unsupervised} in that it requires knowing the number of examples per cluster (partial information). 
We will be comparing with Constrained Clustering Networks~\citep{hsu2017learning,hsu2018multiclass}, a recent state-of-the art learning to cluster method, on their split of Omniglot which we denote Omniglot-CCN.\footnote{We make the choice to apply our method on their task rather than the opposite because their method is much slower and more complicated to run. By solving the same task as them using the same architectures, we can compare directly with the results from their paper.}
Omniglot-CCN is harder than the usual split, because the training and testing splits contain different alphabets. Furthermore, each episode consists of characters of the same alphabet (more fine-grained) and the number of ways varies (20 to 47 characters per set).
We start by training ProtoNet on classic LT few-shot classification with a center loss of $0.1$. At test-time, we embed the set to cluster and cluster the embeddings using Sinkhorn K-Means.
Because the predicted cluster indices are permutation invariant, we run the Hungarian algorithm to find the permutation that maximizes the accuracy. The resulting accuracy is called the \textbf{clustering accuracy}~\citep{hsu2017learning}.
We find that CentroidNet outperform all ``flavors'' of CCN by a margin (86.8\% vs. 83.3\% highest), while being simpler and running about 100 times faster (the data is only embedded once).
We provide additional few-shot clustering results in Section~\ref{app:expfsc} of the appendix.

\paragraph{Transductive Few-Shot Classification  [Table~\ref{table:transductive}].}
In transductive few-shot classification, the learner is allowed to make predictions jointly on the query set. We explore using Sinkhorn K-Means to post-process non-transductive predictions in order to solve transductive few-shot classification. 
We start from the non-transductive predictions given by the ProtoNet and FEAT implementations of~\citet{ye2020fewshot}. Using Sinkhorn K-Means we search for the optimal transport plan between query points and classes. The cost of assigning a query point to a class is taken equal to its predicted negative log-likelihood for that class. We solve transductive few-shot classification on \textit{mini}ImageNet and \textit{tiered}ImageNet and compare the results with TPN~\citep{liu2018learning}, TEAM~\citep{liu2018learning}, and CAN+\citep{hou2019cross}. We achieve the best scores after starting from a strong non-transductive baseline. 
Some people might argue that our comparison is unfair because we explicitly use the uniform distribution of query labels unlike other methods~\citep{vinyals2016matching,liu2018learning,qiao2019transductive}. However, we wish to point out that there is no reason to believe that other methods aren't implicitly leveraging this assumption (even in our case, we could infer the label distribution by treating it as a hyperparameter). 

\begin{table}
\centering
\caption{Transductive 5-way 5-shot classification on \textit{mini}ImageNet and \textit{tiered}ImageNet with the ResNet-12 backbone. ProtoNet numbers are reproduced from the implementation of~\citet{ye2020fewshot}.\label{table:transductive}}
\begin{tabular}{lll}
\toprule
\textbf{Transductive Methods} & \multicolumn{2}{c}{\textbf{Transductive Accuracy}} \\
\textcolor{lg}{dataset $\longrightarrow$}  & \multicolumn{1}{c}{\textit{mini}ImageNet} & \multicolumn{1}{c}{\textit{tiered}ImageNet}\\
\midrule
TPN~\citep{liu2018learning} & 75.65  & - \\
TEAM~\citep{qiao2019transductive} & 75.90  & - \\
ProtoNet~\citep{ye2020fewshot} (non-transductive) & 80.40\ci{0.57} & 84.24\ci{0.65}\\
CAN+~\citep{hou2019cross} & 80.64 & 84.93\\
\ifadditionaltransductive
FEAT~\citep{ye2020fewshot} (non-transductive) & 80.99\ci{0.61} & 84.58\ci{0.63}\\
\fi
ProtoNet+Sinkhorn (ours) & \textbf{82.79}\ci{0.57} & \textbf{86.35}\ci{0.63}\\
\ifadditionaltransductive
FEAT+Sinkhorn (ours) & \textbf{83.54}\ci{0.58} & \textbf{87.47}\ci{0.60}\\
\fi
\bottomrule
\end{tabular}
\end{table}

\section{Conclusion}

Motivated by the apparent lack of diversity of some popular few-shot classification benchmarks, we have proposed a new baseline that attempts to solve them without using support set labels at test-time (LT). 
We find that Omniglot can be solved without LT, and report accuracies on several popular same-domain and cross-domain benchmarks.
By comparing our NLT baseline to state-of-the-art LT methods, we confirm that cross-domain few-shot classification is significantly harder and dependent on using LT.
In general, our NLT accuracies on any combination of dataset and architecture can be taken as a future reference of the baseline performance to be expected from only using a good representation, without any adaptation to test-time labels.
We hope that our work has raised awareness about some limitations of current few-shot learning benchmarks. 
Our results support recent developments of harder multi-domain benchmarks.

\section{Broader Impact}

Our work discusses the way researchers evaluate their methods, in a specific topic of machine learning. We are motivated by the hypothesis that benchmarks which are not representative of real world situations might lead to overconfidence and could result in unpredictable behavior on deployment.

\begin{ack}
We thank Min Lin and Eugene Belilovsky for insightful discussions on few-shot classification. 
We thank Jose Gallego for insightful discussions on Sinkhorn K-Means. 
We thank Pascal Lamblin and Eleni Triantafillou for helping with the Meta-Dataset benchmark. 
This research was partially supported by the NSERC Discovery Grant RGPIN2017-06936, a Google Focused Research Award and the Canada CIFAR AI Chair Program. 
We also thank Google for providing Google Cloud credits.
\end{ack}

\bibliography{l2c-bib}
\bibliographystyle{abbrvnat}

\clearpage

\appendix

\additionaltransductivetrue

\section{Links to the Code}

The code for same-domain experiments on \textit{mini}ImageNet, \textit{tiered}ImageNet, CUB, and cross-domain experiments on \textit{mini}ImageNet$\rightarrow$CUB is forked from the original FEAT code\footnote{\url{https://github.com/Sha-Lab/FEAT}} and is available at \url{https://anonymous.4open.science/r/7fd48c5c-1a56-426d-a980-6bdc1c03f06a/}.

The code for cross-domain experiments on Meta-Dataset is forked from the original Meta-Dataset code\footnote{\url{https://github.com/google-research/meta-dataset}} and is available at
\url{https://anonymous.4open.science/r/cccb838f-8401-44f7-bfed-8da9c8c69aef/}.

The code for few-shot clustering experiments and same-domain experiments on Omniglot is forked from the original ProtoNet implementation\footnote{\url{https://github.com/jakesnell/prototypical-networks}} and is available at \url{https://anonymous.4open.science/r/cbedf355-0158-4cbd-93f0-854d6af33cf4/}.

\section{Additional Related Work from Clustering Literature \label{sec:additionalrelatedwork}}

\textbf{Supervised clustering.}  Supervised clustering is defined in~\citet{finley2005supervised} as ``learning how to cluster future sets of items [...] given sets of items and complete clusterings over these sets''. They use structured SVM to learn a similarity-metric between pairs of items, then run a fixed clustering algorithm which optimizes the sum of similarities of pairs in the same cluster. In follow-up work~\citep{finley2008supervised}, they use K-Means as the clustering algorithm. A main difference with our work is that we learn a nonlinear embedding function, whereas they assume linear embeddings. The work of~\citet{awasthi2010supervised} is also called supervised clustering, although they solve a very different problem. They propose a clustering algorithm which repetitively presents candidate clusterings to a ``teacher'' and actively requests feedback (supervision). 

\textbf{Learning to cluster.} Recent deep learning literature has preferred the term ``learning to cluster'' to ``supervised clustering''. Although the task is still the same, the main difference is the learning of a similarity metric using deep networks. Because of this aspect, these works are often classified as falling in the ``metric learning'' literature. \citet{hsu2017learning,hsu2018multiclass} propose a Constrained Clustering Network (CCN) for learning to cluster based on two distinct steps: learning a similarity metric to predict if two examples are in the same class, and optimizing a neural network to predict cluster assignments which tend to agree with the similarity metric. CCNs obtained the state-of-the-art results when compared against other supervised clustering algorithms, we will thus use CCN as a strong baseline. In our experiments, Centroid Networks improve over CCN on their benchmarks, while being simpler to train and computationally much cheaper.

\textbf{Semi-supervised \& constrained clustering.} Semi-supervised clustering consists of clustering data with some supervision in the form of ``this pair of points should be/not be in the same cluster''. Some methods take the pairwise supervision as hard constraints~\citep{wagstaff2001constrained}, while others (including CCN) learn metrics which tend to satisfy those constraints~\citep{bilenko2004integrating}. See the related work sections in~\citet{finley2005supervised,hsu2017learning}. 

\textbf{Sinkhorn K-Means.} The idea of formulating clustering as minimizing a Wasserstein distance between empirical distributions has been proposed several times in the past~\citep{mi2018variational}. \citet{canas2012learning} explicit some theoretical links between K-Means and the Wasserstein-2 distance. The most similar work to Sinkhorn K-Means is Regularized Wasserstein-Means~\citep{mi2018regularized}, but they use another method for solving optimal transport. Specifically using Sinkhorn distances (regularized Wasserstein distances) for clustering has even been suggested in~\citet{genevay2017learning}.  However, as we could not find an explicit description of the Sinkhorn K-Means anywhere in the literature, we coin the name and explicitly state the algorithm in Section~\ref{sec:sinkhornkmeans}. To our knowledge, we are the first to use Sinkhorn K-Means in the context of learning to cluster and to scale it up to more complex datasets like \miniimagenet{}.  Note that our work should not be confused with Wasserstein K-Means and similar variants, which consist in replacing the squared $L_2$ base-distance in K-Means with a Wasserstein distance.

\section{Additional information on Sinkhorn K-Means.\label{app:sinkhorn}}

\paragraph{Sinkhorn Distances.} The \textit{Wasserstein-2} distance is a distance between two probability masses $p$ and $q$. Given a base distance $d(x,x')$, we define the cost of transporting one unit of mass from $x$ to $x'$ as $d(x,x')^2$. The Wasserstein-2 distance is defined as the cheapest cost for transporting all mass from $p$ to $q$. When the transportation plan is regularized to have large entropy, we obtain Sinkhorn distances, which can be computed very efficiently for discrete distributions~\citep{cuturi2013sinkhorn,cuturi2014fast} (entropy-regularization makes the problem strongly convex). Sinkhorn distances are the basis of the Sinkhorn K-Means algorithm, which is the main component of Centroid Networks.
In Algorithm~\ref{alg:sinkhorn}, we describe the Sinkhorn algorithm in the particular case where we want to transport mass from the weighted data points $(x_i, R_j)$ to the weighted centroids $(c_j, C_j)$, where $R_j$ and $C_j$ are the weights of the data points and centroids, respectively. In practice, we leverage the log-sum-exp trick in computing the Sinkhorn distances to avoid numerical underflows.

\paragraph{Optimization problem} Both of Sinkhorn and Regular K-Means can be formulated as a joint minimization in the centroids $c_j \in \R^d$ (real vectors) and the assignments $p_{i,j} \geq 0$ (scalars) which specify how much of each point $x_i$ is assigned to centroid $c_j$:
\begin{itemize}
\item \textbf{K-Means.} Note that compared to the usual convention, we have normalized assignments $p_{i,j}$ so that they sum up to 1. \vspace{-5mm}
{\small
\begin{align*}
\begin{array}{ll@{}ll}
\text{minimize}  & \displaystyle\min_{p,c} \sum_{i=1}^N \sum_{j=1}^K p_{i,j} ||x_i -& c_j||^2 \\
\text{subject to}& \displaystyle \sum_{j=1}^K p_{i,j} =\frac 1  N,  &i\in 1\!:\!N\\
&                                                p_{i,j} \in \{0,\frac 1 N\}, &i\in 1\!:\!N, \,\, j\in 1\!:\!K\\
\end{array}
\end{align*}
}
\item \textbf{Sinkhorn K-Means.}
{\small
\begin{align*}
\begin{array}{ll@{}ll}
\text{minimize}  & \displaystyle\min_{p,c} \sum_i \sum_j p_{i,j} ||x_i -& c_j||^2 - \gamma \underbrace{H(p)}_{\textrm{entropy}}\\
\text{subject to}& \displaystyle  \sum_{j=1}^K p_{i,j} =\frac 1 N,  &i\in 1\!:\!N\\
& \displaystyle \sum_{i=1}^N p_{i,j} =\frac 1 K,  &j\in 1\!:\!K\\
&                                                p_{i,j} \geq 0 &i\in 1\!:\!N, \,\, j\in 1\!:\!K
\end{array}
\end{align*}
}

where $H(p) = -\sum_{i,j} p_{i,j} \log p_{i,j}$ is the entropy of the assignments, and $\gamma \geq 0 $ is a parameter tuning the entropy penalty term.
\end{itemize}

\paragraph{Differences between Sinkhorn vs. Regular K-Means.} The first difference is that K-Means only allows hard assignments $p_{i,j} \in \lbrace 0,\frac 1 N \rbrace$, that is, each point $x_i$ is assigned to exactly one cluster $c_j$. On the contrary, the Sinkhorn K-Means formulation allows soft assignments $p_{i,j} \in [0,\frac 1 N]$, but with the additional constraint that the clusters have to be balanced, i.e., the same amount of points are soft-assigned to each cluster $\sum_i p_{i,j} = \frac{1}{K}$. 
The second difference is the penalty term $-\gamma H(p)$ which encourages solutions of high-entropy, i.e., points will tend to be assigned more uniformly over clusters, and clusters more uniformly over points. Adding entropy-regularization allows us to compute $p_{i,j}$ very efficiently using the work of~\citet{cuturi2013sinkhorn}. Note that removing the balancing constraint  $\sum_i p_{i,j} = \frac{1}{K}$ in the Sinkhorn K-Means objective would yield a regularized K-Means objective with coordinate update steps identical to EM in a mixture of Gaussians (with $p_{i,j}$ updated using softmax conditionals).

\paragraph{Why is Sinkhorn K-means expected to improve performance ?}
The ablation study in Section~\ref{sec:ablation} shows that using Sinkhorn K-Means instead of K-Means is the most decisive factor in improving performance. There are mainly two possible explanations :
\begin{enumerate}
\item Sinkhorn K-Means is particularly well adapted to the few-shot clustering and unsupervised few-shot classification problems because it strictly enforces the number of images per cluster, whereas K-Means does not.
\item Sinkhorn K-Means is likely to converge better than K-means due to the entropy-regularization factor of the Sinkhorn distance. 
\end{enumerate}
To illustrate the second point, consider the limit case where the regularization factor of Sinkhorn distance goes to infinity ($\gamma\to \infty$). Then, the assignments in Sinkhorn K-Means become uniform (each cluster is assigned equally to all points), and all the centroids converge -- in one step -- to the average of all the points, reaching global minimum. This is by no means a rigorous proof, but the limit case suggests that Sinkhorn K-Means converges well for large enough~$\gamma$. This behavior is to be contrasted with K-means, for which convergence is well known to depend largely on the initialization.

\paragraph{What is the effect of using weighted vs. unweighted averages ? }
One could argue that comparing CentroidNets with ProtoNets is unfair because using Sinkhorn K-Means leads to centroids which are weighted averages, whereas ProtoNet prototypes are restricted to be unweighted averages. Therefore, we run Centroid Networks on \textit{mini}Imagenet, but under the constraint that centroids to be unweighted averages of the data points. To do so, starting from the soft weights, we reassign each data point only to its closest centroid, and compute the unweighted averages. The comparison between ProtoNets and CentroidNets is now fair in the sense that both prototypes and centroids use unweighted averages. (Numbers below based on official ProtoNet implementation~\citep{snell2017prototypical})
\begin{itemize}
\item Unsupervised accuracy on \textit{mini}Imagenet is $0.5508 \pm 0.0072$ for weighted average and $0.5497 \pm 0.0072$ for unweighted average. The difference is not significant.
\item Clustering accuracy on \textit{mini}Imagenet is $0.6421 \pm 0.0069$ for weighted average and $0.6417 \pm 0.0069$ for unweighted average. The difference is also not significant.
\end{itemize}
This experiment suggests that using weighted averages does not bring an unfair advantage, and therefore does not invalidate our comparison. More generally, instead of trying to tune ProtoNets and CentroidNets as well as possible, we try to make ProtoNets and CentroidNets more comparable by using the same architectures and representation.

\section{Implementation Details\label{sec:implementationdetails}}

\paragraph{Splits.} For Omniglot which has 1623 classes of handwritten characters, we consider the ``Vinyals'' splits~\citep{vinyals2016matching}. For \textit{mini}ImageNet which has 100 object classes, we consider the ``Ravi'' splits~\citep{ravi2016optimization} of 64 training, 16 validation, 20 testing classes. For CUB which has 200 bird species, we use the same split as~\citet{ye2020fewshot} which consists of 100 training, 50 validation, and 50 testing classes. For Meta-Dataset, we consider the official splits and sampling scheme, which features variable numbers of ways and shots~\citep{triantafillou2019meta}. For Omniglot-CCN, we consider the same splits as\citet{hsu2017learning} : 30 background alphabets are used for training and 20 evaluation alphabets are used for validation (there is no testing set).

\paragraph{Backbones.} We consider two backbones throughout the experiments: the Conv-4 classically used in few-shot learning~\citep{snell2017prototypical,finn2017model,vinyals2016matching} and the ResNet-12~\citep{ye2020fewshot,lee2019meta}. We also consider the CCN architecture~\citep{hsu2017learning,hsu2018multiclass,hsu2018unsupervised} for the few-shot clustering experiments, and the ResNet-18 architecture for Meta-Dataset~\citep{triantafillou2019meta}.

\paragraph{Reference implementations.} All our CentroidNet implementations are derived from specific reference ProtoNet implementations~\citep{snell2017prototypical,triantafillou2019meta,ye2020fewshot}. Unless otherwise specified, we always use exactly the same training procedures and hyperparameters as the reference implementations. We have grouped our experiments below based on their reference implementations.

\paragraph{Code based on~\citet{ye2020fewshot}.} 
For same-domain experiments on \textit{mini}ImageNet, \textit{tiered}ImageNet, CUB (Section~\ref{sec:expcrosstask}), cross-domain experiments on \textit{mini}ImageNet$\rightarrow$CUB (Section~\ref{sec:expcrossdomain}), and transductive experiments on \textit{mini}ImageNet, \textit{tiered}ImageNet (Section~\ref{sec:expother}), we derive CentroidNet from the implementation of~\citet{ye2020fewshot} using the Conv-4 and ResNet-12 architecture.
We denote ProtoNet*~\citep{ye2020fewshot} the accuracies reported from their paper and ProtoNet* (repro) the accuracies reproduced from their code.\footnote{\url{https://github.com/Sha-Lab/FEAT}} 
We adopt exactly the same training strategy for ProtoNet and CentroidNet using the default hyperparameters of their implementation. 
We start from the provided pretrained checkpoints, which are given by training a classifier on all training classes (non-episodic training) with data augmentation.
Then, we finetune the models on 5-way 5-shot LT few-shot classification episodes using Adam optimizer with initial learning rates of $0.0001$ for Conv-4 and $0.0002$ for ResNet-12.
We do not use a center loss, and use a temperature of 32 for Conv-4 and 64 for ResNet-12 to rescale the squared Euclidean distances between prototypes/centroids and data embeddings before feeding them to the  softmax layer/Sinkhorn K-Means.
We run a hyperparameter search over Sinkhorn regularization values $0.03,0.1,0.3,1,3,10,30$ (values given for squared Euclidean distances rescaled by the temperature). The optimal value is always $\gamma=3$ except for CUB ($\gamma=1$), \textit{mini}ImageNet$\rightarrow$CUB ($\gamma=1$), \textit{mini}ImageNet with RestNet-12 ($\gamma=10$), and transductive \textit{tiered}ImageNet ($\gamma=1$, value given for Sinkhorn on logit values).

\paragraph{Code based on~\citet{snell2017prototypical}.} For same-domain experiments on Omniglot (Section~\ref{sec:expcrosstask}) and the ablation study (Section~\ref{sec:ablation}), we use the Conv-4 architecture from the official Prototypical Networks~\citep{snell2017prototypical} implementation.\footnote{\url{https://github.com/jakesnell/prototypical-networks}} This results in a 64-dimensional embedding for Omniglot and 1600-dimensional embedding for \miniimagenet{}. For \miniimagenet{}, we pretrain the embedding function using prototypical networks to solve 30-way problems instead of 5, which is the recommended trick in the paper~\citep{snell2017prototypical}. For Omniglot, we train from scratch on 5-way 5-shot and 20-way 5-shot episodes (number of ways for training and testing match). We use a center loss of $1$ and Sinkhorn regularization of $\gamma=1$, on unnormalized squared Euclidean distances. 

\paragraph{Code based on~\citet{snell2017prototypical} and~\citet{hsu2017learning}} For few-shot clustering (Section~\ref{sec:expother}), we start from the official ProtoNet implementation and train both Conv-4 and CCN architectures on the CCN splits of Omniglot. We use a center loss of $0.1$ and Sinkhorn regularization of $\gamma=1$, on unnormalized squared Euclidean distances.

\paragraph{Code based on~\citet{triantafillou2019meta}} For cross-domain experiments on Meta-Dataset (Section~\ref{sec:expcrossdomain}) we derive CentroidNet from their implementation of ProtoNet\footnote{\url{https://github.com/google-research/meta-dataset}} which uses the ResNet-18 architecture. We train with a center loss of $0.01$ and Sinkhorn regularization $\gamma=0.1$ after rescaling the squared Euclidean distances by the number of dimensions.

\section{Additional Few-Shot Clustering Results\label{app:expfsc}}

\begin{table}[h]
\caption{Clustering accuracies for Centroid Networks and K-Means (raw and ProtoNet features) on Omniglot 5-way 5-shot, Omniglot 20-way 5-shot, and \textit{mini}ImageNet 5-way 5-shot. \label{table:fsc}}
\centering
\begin{tabular}{l l l l l}
\toprule
 \textbf{FSC Method} & \multicolumn{4}{c}{\textbf{Clustering Accuracy} } \\
\textcolor{lg}{dataset$\rightarrow$}  & \textbf{Omniglot-5} & \textbf{Omniglot-20} & \textbf{\textit{mini}ImageNet-5} & \textbf{\textit{tiered}ImageNet-5} \\
\midrule
K-Means (Raw) & 45.2 {\color{lg} $\pm$ 0.5} & 30.7  {\color{lg} $\pm$ 0.2} & 41.4  {\color{lg} $\pm$ 0.4} & -\\ 
K-Means (Conv) & 83.5  {\color{lg}$\pm$ 0.8} & 76.8  {\color{lg}$\pm$ 0.4} & 48.7  {\color{lg}$\pm$ 0.5} & -\\
CentroidNet (Conv) & \textbf{99.6}  {\color{lg}$\pm$ 0.1}  & \textbf{99.1}  {\color{lg}$\pm$ 0.1} & \textbf{64.5}  {\color{lg}$\pm$ 0.7} & - \\
CentroidNet (ResNet) & -  & - & 77.3  & 82.49\\
\bottomrule
\end{tabular}
\end{table}

\clearpage

\section{Ablation Study\label{sec:ablation}}

\begin{figure*}[h]
\includegraphics[width=\textwidth]{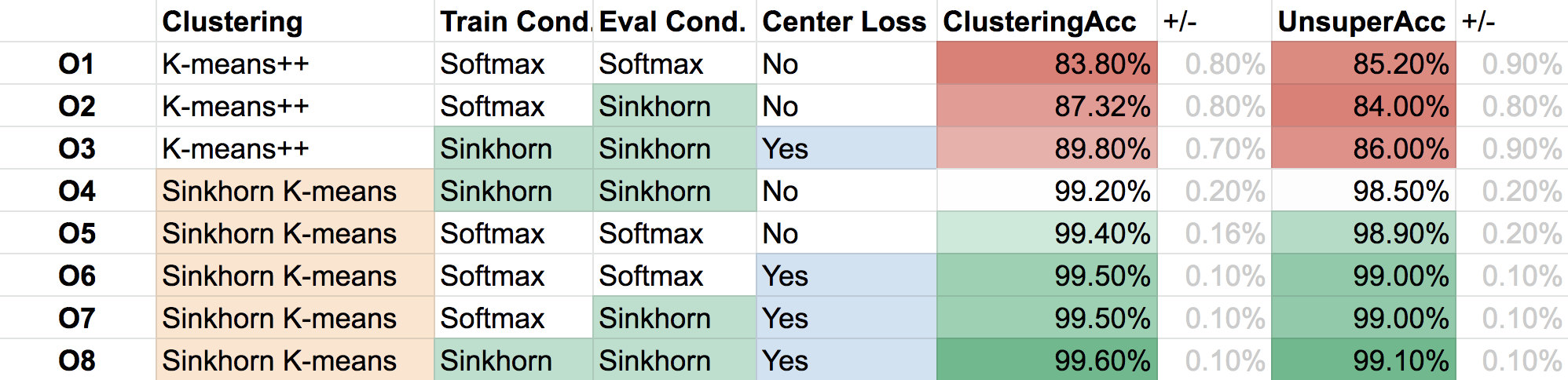}
\caption{Omniglot 5-way 5-shot Ablation Study\label{fig:ablationomniglot}}
\end{figure*}

\begin{figure*}[h]
\includegraphics[width=\textwidth]{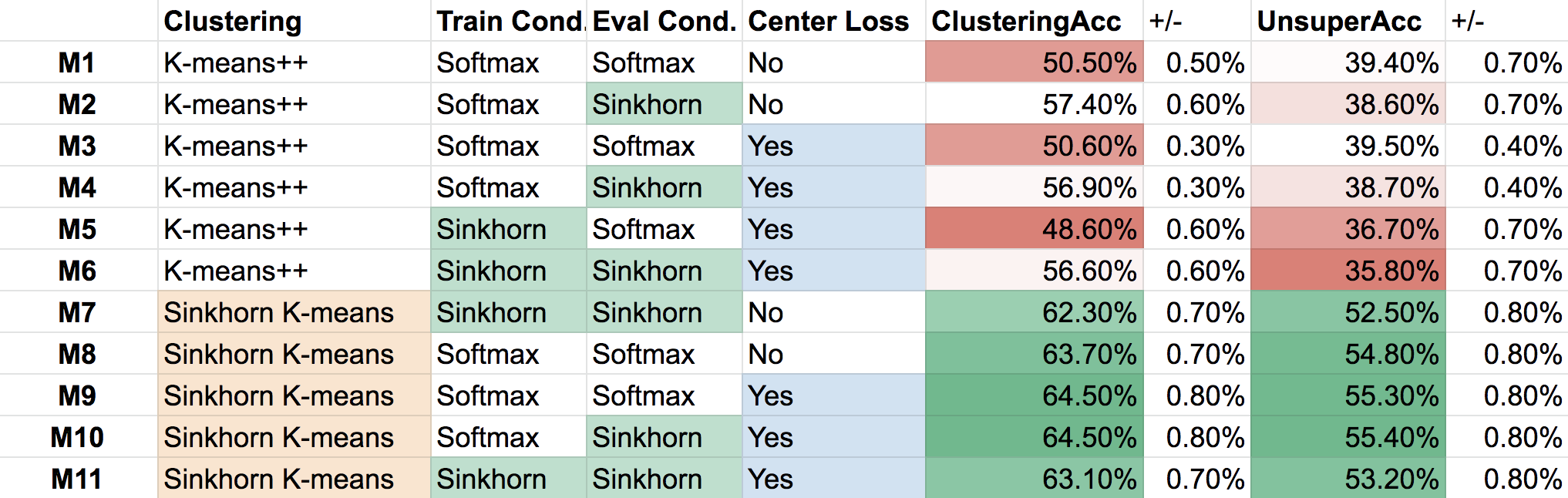}
\caption{\textit{mini}ImageNet 5-way 5-shot Ablation Study\label{fig:ablationminiimagenet}}
\end{figure*}

\textbf{[Figures~\ref{fig:ablationomniglot},\ref{fig:ablationminiimagenet}]} We conduct an ablation study on Omniglot (5-way 5-shot) and \textit{miniImageNet} (5-way 5-shot) to determine the effect and importance of the various proposed tricks and components. Implementations in this section are based on the official ProtoNet implementation~\citep{snell2017prototypical}, thus numbers might differ slightly from the main paper.
\begin{itemize}
\item \textbf{K-Means vs. Sinkhorn K-Means.} From comparing O3 to O4, O1 to O5, M6 to M7, M1 to M8, it appears that using Sinkhorn K-Means instead of K-Means++ is the most beneficial and important factor.
\item \textbf{Center Loss}. From comparing O2 to O3, O5 to O6, O4 to O8, M7 to M11, M8 to M9, center loss seems to be beneficial (although the significance is at the limit of the confidence intervals). It is the second most influential factor.
\item \textbf{Softmax vs. Sinkhorn conditionals} (at train and test time). For training, it is not clear whether using Sinkhorn or Softmax conditionals is beneficial or not. For evaluation, from comparing M1 to M2, M3 to M4, M5 to M6, it seems that Sinkhorn conditionals are better if the metric is clustering accuracy, while Softmax conditionals might be better if the metric is unsupervised accuracy, although the effect seems to be negligible (see how the color patterns are inverted).
\end{itemize}

\clearpage

\section{Confidence Intervals\label{sec:expmetadatasetci}}

We give 95\% confidence intervals for the accuracies reported in the experimental section.

\intervalstrue

\begin{table}[H]
    \centering
    \caption{Confidence Intervals for Same-Domain Benchmarks}
    
\end{table}

\begin{table}[H]
    \centering
    \caption{Confidence Intervals for Cross-Domain Benchmarks}
    
\end{table}

\end{document}